# Modification of the Elite Ant System in Order to Avoid Local Optimum Points in the Traveling Salesman Problem

*Majid Yousefikhoshbakht\*, Department of Mathematics and Computer Science, Amirkabir University of Technology, Tehran, Iran*

*Farzad Didehvar, Department of Mathematics and Computer Science, Amirkabir University of Technology, Tehran, Iran*

*Farhad Rahmati, Associate Professor, Department of Mathematics and Computer Science, Amirkabir University of Technology, Tehran, Iran*

**Abstract**

This article presents a new algorithm which is a modified version of the elite ant system (EAS) algorithm. The new version utilizes an effective criterion for escaping from the local optimum points. In contrast to the classical EAC algorithms, the proposed algorithm uses only a global updating, which will increase pheromone on the edges of the best (i.e. the shortest) route and will at the same time decrease the amount of pheromone on the edges of the worst (i.e. the longest) route. In order to assess the efficiency of the new algorithm, some standard traveling salesman problems (TSPs) were studied and their results were compared with classical EAC and other well-known meta-heuristic algorithms. The results indicate that the proposed algorithm has been able to improve the efficiency of the algorithms in all instances and it is competitive with other algorithms.

## 1. Introduction

Ant colony optimization (ACO) which has been inspired by the behavior of real ants seeking a path between their colony and a source of food is one of the most important meta-heuristic algorithms. Initially proposed by Marco Dorigo in 1992, the first algorithm which was called Ant System (AS) aimed at searching for an optimal path between two nodes in a graph. Therefore, a problem is divided into some sub-problems in which the simulated ants are expected to select the next node based on the amount of the pheromone in a trail and the distance to the next node. The decision for choosing the unvisited $N_i$ node by ant k located in node *i* is made based on formula (1) where $\tau_{ij}$ indicates the amount of pheromone on (i, j) edge while $\eta_{ij}$ shows inverse distance between i and j. However, both are powered by $\alpha$ and $\beta$ which can be changed by the user. Therefore, their relative importance can be altered.

$$P_{ij}^k = \begin{cases} \dfrac{\tau_{ij}^{\alpha} \eta_{ij}^{\beta}}{\sum_{j \in N_i} \tau_{ij}^{\alpha} \eta_{ij}^{\beta}} & if \quad j \in N_i \\ 0 & if \quad j \notin N_i \end{cases} \qquad (1)$$

Ants release $\Delta\tau_{ij}$ which is called "pheromone information" on the respective path while moving from node *i* to node *j*. $\Delta\tau_{ij}$ can be calculated by formula (2).

$$\tau_{ij}(t) \leftarrow \tau_{ij}(t) + \Delta\tau_{ij} \qquad (2)$$

Moreover, the algorithm like its natural version makes use of pheromone evaporation in order to prevent rapid convergence of ants to a sub-optimal path. In other words, pheromone density is reduced in each iteration by $0 \leq \rho \leq 1$ ( set by the user ). If $\tau$ is the matrix for the existing pheromone on the edges of the respective graph, then it is updated in each iteration by the formula (3).

$$\tau \leftarrow (1 - \rho)\tau \qquad \rho \in [0,1] \qquad (3)$$

**Refrences**